\documentclass[journal]{IEEEtran}

\usepackage{graphicx}
\usepackage{amsmath}
\usepackage{amsfonts, mathtools}
\usepackage{hyperref}
\usepackage{multirow}
\usepackage{caption}
\usepackage{subcaption}
\usepackage{graphicx}
\usepackage[export]{adjustbox}
\usepackage{xcolor}
\usepackage{soul}
\usepackage{cite}

\usepackage{algorithm}
\usepackage{algpseudocode}
\algnewcommand\algorithmicinput{\textbf{Input:}}
\algnewcommand\Input{\item[\algorithmicinput]}
\algnewcommand\algorithmicoutput{\textbf{Output:}}
\algnewcommand\Output{\item[\algorithmicoutput]}

\newcommand{\tmath}[1]{\texorpdfstring{#1}{TEXT}}
\newcommand{\take}[1]{\textit{#1}}
\newcommand{\settings}[1]{(Settings: #1)}
\newcommand{\heading}[1]{\noindent \textbf{#1} \space}
\newcommand{\headingTwo}[1]{\noindent \textit{#1} \space}

\def\BibTeX{{\rm B\kern-.05em{\sc i\kern-.025em b}\kern-.08em
    T\kern-.1667em\lower.7ex\hbox{E}\kern-.125emX}}

\begin{document}

\title{Adv-MIA: Efficient Membership Inference Attacks on DNNs using Adversarial Perturbations}
\title{Membership Inference Attacks on DNNs using Adversarial Perturbations}


\author{Hassan~Ali$^1$, Adnan~Qayyum$^1$, Ala~Al-Fuqaha$^2$, Junaid~Qadir$^3$
\thanks{$^1$IHSAN Lab, Information Technology University, Lahore, Pakistan (hassanalikhatim@gmail.com, adnan.qayyum@itu.edu.pk)\\ $^2$Hamad Bin Khalifa University, Qatar (aalfuqaha@hbku.edu.qa).\\ $^3$Corresponding author (jqadir@qu.edu.qa). Qatar University, Doha, Qatar.
}}






\maketitle
\begin{abstract}
Several membership inference (MI) attacks have been proposed to audit a target DNN. Given a set of subjects, MI attacks tell which subjects the target DNN has seen during training. This work focuses on the post-training MI attacks emphasizing high confidence membership detection---True Positive Rates (TPR) at low False Positive Rates (FPR). Current works in this category---likelihood ratio attack (LiRA) and enhanced MI attack (EMIA)---only perform well on complex datasets (e.g., CIFAR-10 and Imagenet) where the target DNN overfits its train set, but perform poorly on simpler datasets (0\% TPR by both attacks on Fashion-MNIST, 2\% and 0\% TPR respectively by LiRA and EMIA on MNIST at 1\% FPR). 
To address this, firstly, we unify current MI attacks by presenting a framework divided into three stages---preparation, indication and decision.
Secondly, we utilize the framework to propose two novel attacks: (1) Adversarial Membership Inference Attack (AMIA) efficiently utilizes the membership and the non-membership information of the subjects while adversarially minimizing a novel loss function, achieving 6\% TPR on both Fashion-MNIST and MNIST datasets; and (2) Enhanced AMIA (E-AMIA) combines EMIA and AMIA to achieve 8\% and 4\% TPRs on Fashion-MNIST and MNIST datasets respectively, at 1\% FPR. 
Thirdly, we introduce two novel augmented indicators that positively leverage the loss information in the Gaussian neighborhood of a subject. This improves TPR of all four attacks on average by 2.5\% and 0.25\% respectively on Fashion-MNIST and MNIST datasets at 1\% FPR.
Finally, we propose simple, yet novel, evaluation metric, the running TPR average (RTA) at a given FPR, that better distinguishes different MI attacks in the low FPR region. 
We also show that AMIA and E-AMIA are more transferable to the unknown DNNs (other than the target DNN) and are more robust to DP-SGD training as compared to LiRA and EMIA.
\end{abstract}


\section{Introduction}

Deep Learning (DL) algorithms, particularly Deep Neural Networks (DNNs), are now being used for decision-making in several sensitive domains, such as IoT devices, healthcare~\cite{qayyum2020secure}, and purchase recommendations~\cite{shokri2017membership}, where an individual's data privacy is the first and the foremost concern~\cite{carlini2022membership, ye2022enhanced, khalid2023privacy}. However, several previous works have shown that DNNs can leak sensitive information about their training data. Privacy attacks~\cite{tramer2022truth, boenisch2023federated}, most notably membership inference (MI) attacks~\cite{shokri2017membership, carlini2022membership, ye2022enhanced, nguyen2023active, jagielski2023students}, exploit this leakage to quantify the privacy of a target DNN. Given a trained target DNN and a subject ($x$:input to DNN, $y$:ground truth), an MI attack infers the training membership of the subject---whether or not the target DNN has seen the subject during training. This work focuses on the inference time MI attacks on the target DNN with emphasis on high confidence membership detection---True Positive Rates (TPR) at low False Positive Rates (FPR) (see Section~\ref{sec:setup_metrics} for definitions).

Carlini et al. \cite{carlini2022membership} note that most of the previous attacks are only good at inferring the \textit{non-membership} of a subject, and therefore, should instead be called \textit{non-membership inference} (\textit{non}-MI) attacks. non-MI attacks are not suitable for the threat setting where it is desired to infer the membership of a subject with high confidence. Therefore, Carlini et al. \cite{carlini2022membership} propose the Likelihood Ratio Attack (LiRA) in two variations---online LiRA (n-LiRA) and offline LiRA (f-LiRA)---along with a novel quantitative metric---TPR-FPR curve (in log scale)---to outperform previous attacks in membership inference, specifically in the low FPR region.

However, n-LiRA is computationally inefficient---given $k$ subjects, the membership of which is to be inferred, n-LiRA requires training of $2kN$ shadow DNNs, where $N$ and $k$ are typically in hundreds~\cite{carlini2022membership}---making it almost impossible for the future researchers to reproduce the attack, and for practitioners to audit a target DNN in limited computational settings, which is often the only feasible option. f-LiRA is computationally more efficient---requiring $N$ shadow DNNs to be trained for $k$ subjects---but notably less effective~\cite{carlini2022membership}.
To address this, Ye et al.~\cite{ye2022enhanced} propose Enhanced Membership Inference Attack (EMIA) that uses soft labels from the target DNN to train the shadow DNNs. EMIA performs best in the high FPR region and comparably to n-LiRA in the low FPR region while being significantly computationally efficient---requiring $N$ shadow DNNs to be trained for $k$ subjects.

\vspace{1mm}
\heading{Limitations and Challenges:}
Despite being effective on complex CIFAR-10, CIFAR-100, and Imagenet datasets, both f-LiRA and EMIA do not perform well on MNIST~\cite{ye2022enhanced} and Fashion-MNIST datasets (Section~\ref{sec:results_evaluation}). One possible reason is that the experimental setup of MI attacks only allows training the target DNN on half of the training set that causes the DNN to overfit, yielding a large train-test accuracy gap for complex datasets~\cite{carlini2022membership, ye2022enhanced}. The overfitted target DNN leaks private information to even relatively weaker MI attacks. This is a major limitation, as in practical scenarios, the target DNN is less likely to overfit its training set. Previous works have also observed the overfitting of the target DNN~\cite{carlini2022membership, ye2022enhanced, yeom2018privacy}, but neither of them specifically relates it to MI attacks under-performing on MNIST and Fashion-MNIST, while being effective on complex datasets. We hypothesize two main reasons for this limitation detailed below.

Both f-LiRA and EMIA only leverage the \textit{non-membership information} of the subject (how does the absence of the subject from the training set affect the behavior of the target DNN?) while ignoring its \textit{membership information} (how does the inclusion of the subject in the training set affect the behavior of the target DNN?) in order to gain the computational advantage. As noted by Carlini et al.~\cite{carlini2022membership}, the membership information plays a key role in inferring the membership of the subjects that are \textit{hard-to-fit}.

Both f-LiRA and EMIA use \textit{vanilla membership indicators} that examine the loss of the target DNN precisely over the subjects without considering the loss landscape in the neighborhood (e.g., the Gaussian neighborhood) of the subjects.  However, local loss landscape around the subject might be helpful for membership inference~\cite{jalalzai2022membership}, highlighting the need for \textit{augmented membership indicators}. However, Carlini et al.~\cite{carlini2022membership} observe that simply ensembling the loss in the Gaussian neighborhood of the subject does not improve MI attacks' performance.

These limitations pose the following research challenges:
\begin{enumerate}
    \item How to utilize the membership information of the subjects while being computationally efficient? \label{challenge:efficiecny}
    \item How to positively leverage the loss information in the Gaussian neighborhood of a given subject? \label{challenge:proximity}
\end{enumerate}

\vspace{1mm}
\heading{Findings and Contributions:} In this paper, we first present a unified framework based on the working of current MI attacks by dividing their algorithms into three stages. The \textit{preparation stage} prepares the variables $\mathcal{V}$ that may include transformation functions or shadow DNNs used to guide the subsequent stages. The \textit{indication stage} defines a membership indicator $\mathcal{I}$ that, guided by $\mathcal{V}$, indicates the likelihood of a subject being the training set member. The \textit{decision stage} applies a threshold $\tau$ to $\mathcal{I}$ to finally decide whether the subject is a member or not. For example, LiRA and EMIA respectively train $2kN$ and $N$ shadow DNNs as $\mathcal{V}$, and use the likelihood ratio as $\mathcal{I}$.

Building upon the proposed framework, we contribute towards a better $\mathcal{V}$ by developing an Adversarial Membership Inference Attack (AMIA) that utilizes both, the membership and the non-membership information of a subject while being computationally efficient (only requiring $2N$ shadow DNNs to be trained for $k$ subjects). Given a target DNN and a set of $k$ subjects, AMIA first trains $2N$ shadow DNNs, where each shadow DNN is trained with $k/2$ randomly selected subjects augmented with the original training set as a batch for computational efficiency (Challenge~\ref{challenge:efficiecny}), thereby, holding the (batch-wise) membership information of these $k/2$ subjects and the non-membership information of the remaining $k/2$ subjects. AMIA then adversarially computes small magnitude perturbations to the input of each subject that maximizes the difference between the loss of the member and the non-member shadow DNNs. This lets AMIA exploit the loss landscape of the shadow DNNs in the local neighborhood of the subject (Challenge~\ref{challenge:proximity}) optimally for the membership inference purpose. AMIA notably outperforms both f-LiRA and EMIA on Fashion-MNIST and MNIST datasets and performs on par with f-LiRA (while outperforming EMIA) on the CIFAR-10 dataset.

We also propose Enhanced AMIA (E-AMIA) that exploits soft labels of EMIA to train the shadow DNNs of AMIA. E-AMIA shows comparable performance to AMIA on Fashion-MNIST and MNIST datasets, while notably outperforming others on the CIFAR-10 dataset.

We contribute towards a better $\mathcal{I}$ by extending the likelihood ratio, a vanilla indicator proposed by Carlini et al.~\cite{carlini2022membership}, to define two augmented indicators that exploit the loss landscape in the Gaussian neighborhood of a subject to infer its membership (Challenge~\ref{challenge:proximity}). Our proposed augmented indicators notably improve the performance of all the MI attacks considered in this paper, i.e., the previously proposed attacks (f-LiRA and EMIA) and the newly proposed attacks (AMIA and E-AMIA).

We also present a better evaluation metric of MI attacks in the low FPR region. More specifically, we find that the running average of TPR (RTA) at a given FPR (in log scale) better distinguishes between stronger and weaker attacks in the low FPR region as compared to the TPR at a given FPR (in log scale) proposed by Carlini et al.~\cite{carlini2022membership}.

Finally, we study how well do the variables $\mathcal{V}$ prepared for $k$ subjects by an MI attack for the target DNN transfer to the unknown DNNs trained on the same dataset. We find that EMIA is less transferable to the unknown DNNs as compared to f-LiRA even in the high FPR region where EMIA outperfroms others on the target DNN. This is because EMIA uses soft labels generated by the target DNN which makes it more customized to the target DNN. Interestingly, our proposed AMIA and E-AMIA, in addition to performing better than LiRA and EMIA on the target DNN, transfer well to the unknown DNNs. We attribute this to the transferability property of the adversarial perturbations used by AMIA and E-AMIA.

Our findings and contributions are summarized below.

\begin{enumerate}
    \item We unify MI attacks in the literature under a framework divided into three stages---preparation, indication, and decision stages (which are formalized in Section~\ref{sec:mia}).
    \item We propose a compute-efficient methodology that leverages subject membership information and adversarial perturbations to effectively distinguish members from non-members on MNIST and Fashion-MNIST datasets.
    \item We propose improvements over the likelihood ratio indicator (used by both f-LiRA and EMIA) by leveraging the local loss information in the Gaussian neighborhood of the subject. More specifically, we perturb each subject with the Gaussian noise and compute the likelihood ratio of each perturbed sample separately, which notably improves the effectiveness of the attack.
    \item We study the transferability of MI attacks by using $\mathcal{V}$ prepared for a target DNN to perform membership inference of the subjects on unknown DNNs. Our results indicate that AMIA is the most transferable, respectively followed by E-AMIA and f-LiRA (which give comparable performance) while EMIA is the least transferable.
\end{enumerate}

The source code will be made available at: \href{https://github.com/hassanalikhatim/AMIA}{https://github.com/hassanalikhatim/AMIA}.



\section{Related Work}
Recent works have extensively highlighted several limitations of DNNs in real-world safety-critical scenarios. For example, DNNs are vulnerable to explainability attacks~\cite{ali2022tamp}, adversarial attacks~\cite{khalid2020fadec, ali2021all}, bias~\cite{butt2023towards}, data imperfections~\cite{latif2019caveat} and privacy attacks~\cite{carlini2022membership, khalid2023privacy}. In this section, we present a brief review of research on the MI attack (the most popular type of privacy attacks) and their connection to the adversarial perturbations.

\subsection{Membership Inference Attacks on DNNs}
MI attacks on DNNs can be mainly divided into two categories. \textit{Training time} MI attacks assume that an attacker is capable of influencing the DNN during training, in addition to querying the DNN at test time~\cite{zhang2023agrevader, chen2022amplifying}. For example, Tramer et al.~\cite{tramer2022truth} show that by carefully poisoning a DNN during training can cause the DNN to leak greater membership information at test time. Training-time MI attacks are most relevant to the frameworks such as federated learning, where a DNN is trained on broad data collected through untrustworthy sources. \textit{Inference time} MI attacks assume that the attacker can only query the DNN at test-time without influencing it during the training phase~\cite{jagielski2023combine, carlini2022membership, ye2022enhanced, tan2023blessing, rezaei2022accuracy}. Therefore, inference time MI attacks are relatively more practical because of the limited attacker capabilities. Inference time MI attacks can further be divided into two categories. Attacks that estimate the behavior of the DNNs on given subjects by training several shadow DNNs~\cite{shokri2017membership, carlini2022membership, ye2022enhanced, liu2022membership} are computationally expensive but notably stronger than the attacks that do not train the shadow DNNs but rely on other membership indicator functions~\cite{del2022leveraging, jalalzai2022membership}.

Our work in this paper falls under the category of \textit{inference-time} MI attacks on \textit{black-box} target DNN classifier by \textit{training shadow DNNs} to estimate the behavior of the target DNN. To the best of our knowledge, LiRA~\cite{carlini2022membership} and EMIA~\cite{ye2022enhanced} are two of the most popular and strongest MI attacks in this category. Later works on MI attacks have either largely focused on specific scenarios and applications (network pruning~\cite{yuan2022membership}, multiple models~\cite{jagielski2023combine}, federated learning~\cite{liu2022membership}, distillation~\cite{jagielski2023students}) or fall under a different category of attacks.
Here we briefly discuss both attacks and provide algorithmic details in Section~\ref{sec:preliminaries} after presenting a unified framework of MI attacks.
LiRA has two variations. Online LiRA trains thousands of member (subjects included in the training set) and non-member (subjects not included in the training set) shadow DNNs and learns the distribution of the $\phi$-processed confidences (defined later in eq-\eqref{eq:phi}) of member and non-member shadow DNNs on the given subjects. Online LiRA then computes the likelihood of the $\phi$-processed confidences of the target DNN on the given subjects to fall within the member or the non-member distribution learned previously. Finally, online LiRA uses thresholding to label each subject as a member or a non-member.
Offline LiRA works the same as online LiRA but does not train the non-member shadow DNNs.
EMIA follows a similar algorithm as offline LiRA, but uses soft labels generated by the target DNN to train the non-member shadow DNNs.

\textit{Countermeasures against MI attacks:} Two of the most popular defenses against MI attacks in the current literature are differentially private stochastic gradient descent (DP-SGD) training and $L_2$ regularization~\cite{choquette2021label}. We analyze the effects of both of these approaches on the privacy leakage of the target DNN.


\subsection{Adversarial perturbations in MI attacks}
DNNs have been shown to be vulnerable to adversarial perturbations---small magnitude perturbations carefully crafted through iterative optimization to achieve a targeted goal. The goal of standard adversarial attacks is to change the output of a target DNN $f$ to some target $t$. Formally,
\begin{equation}
    \delta = \underset{\delta}{\operatorname{argmin}}\ (f(x+\delta) = t)
    \label{eq:adversarial_perturabtion_goal}
\end{equation}

Several algorithms have been proposed to achieve the goal in eq-\eqref{eq:adversarial_perturabtion_goal} under numerous threat models resulting in a range of adversarial attacks~\cite{carlini2017towards, madry2018towards, khalid2020fadec, croce2020reliable, ali2023consistent} and countermeasures~\cite{cohen2019certified, ali2023detect, nie2022diffusion, carlini2022certified}.

To the best of our knowledge, only two recent works have leveraged the power of adversarial perturbations to perform MI attacks on DNNs. Del et al.~\cite{del2022leveraging} use the magnitude of adversarial perturbations computed for the target DNN to estimate the likelihood of the membership. Jalalzai et al.~\cite{jalalzai2022membership} use the loss values of the target DNN along the adversarial path computed using multiple perturbation magnitudes to estimate the likelihood of the membership. However, both works aim to compute a general threshold, not specifically customized to the subjects (following Attack S methodology by Ye et al.~\cite{ye2022enhanced}). Additionally, both of these techniques compute adversarial perturbations over the target DNN, assuming a white-box access to compute the gradients, and perform membership inference based on how adversarial perturbations affect the behavior of the target DNN on the subject. In contrast, we introduce a novel objective function well-suited to the MI attacks framework (instead of conventional adversarial attacks objective used by prior works~\cite{del2022leveraging, jalalzai2022membership}) and adversarially optimize the novel objective function over the shadow DNNs (instead of over the target DNN), thereby using both the membership and the non-membership information of the subject, and only use the output probabilities of the target DNN to perform membership inference.

\begin{figure*}
    \centering
    \includegraphics[width=0.8\linewidth]{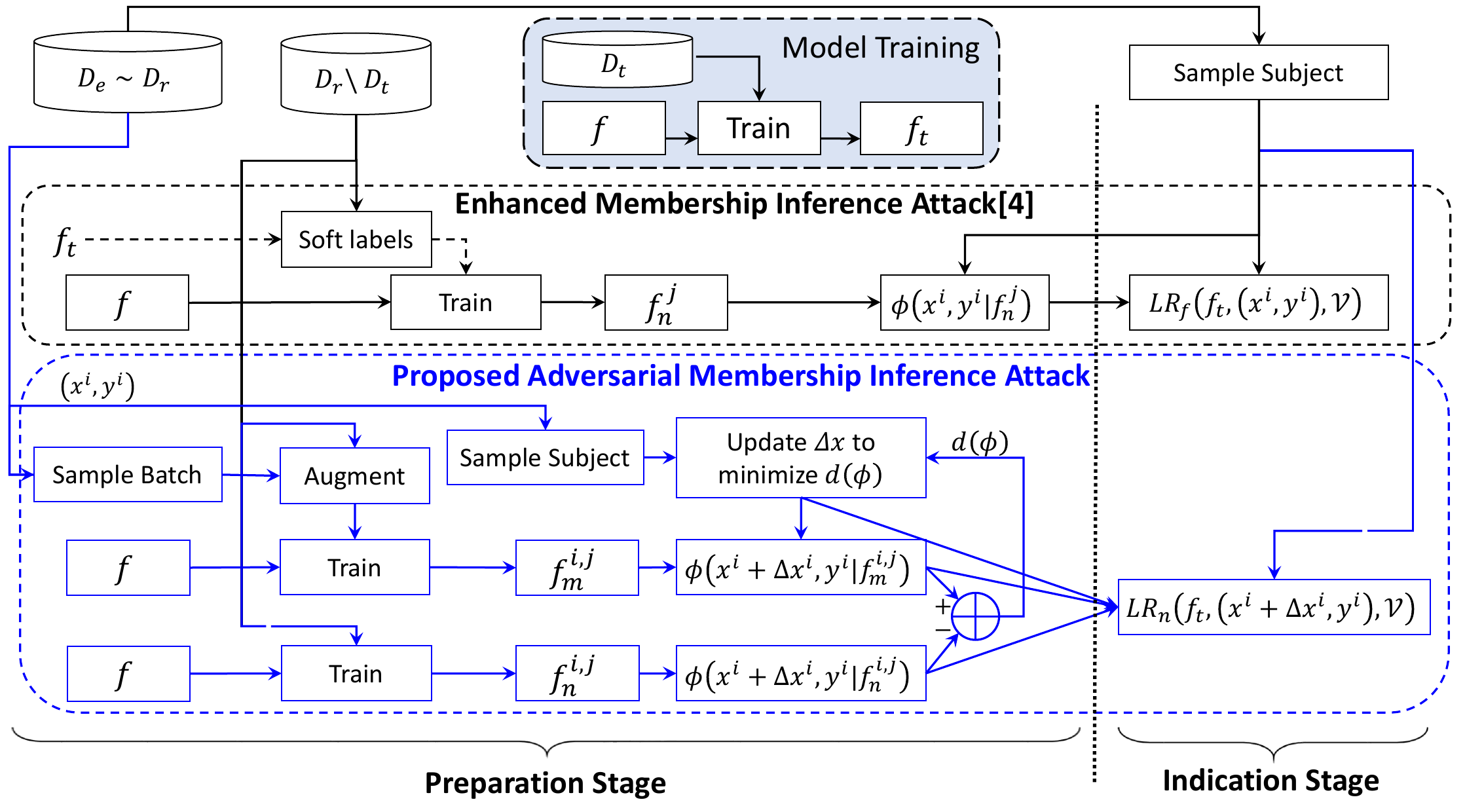}
    \caption{Illustrating and comparing the flow diagram of the preparation and indication stages of EMIA~\cite{ye2022enhanced} and AMIA. Unlike EMIA, which only trains non-member shadow DNNs $f^j_n$, AMIA trains member shadow DNNs $f^{i,j}_m$, along with the non-member shadow DNNs $f^{i,j}_n$ by including subject batches in the training dataset, and optimizes a perturbation for each subject that favors $f^{i,j}_m$ over $f^{i,j}_n$, where $j \in [0..N]$. $\Delta x^i$, $f^j_m$, and $f^j_n$ constitute $\mathcal{V}$ of AMIA, which is then used in the indication stage. \take{Note: The flow diagram of offline LiRA is the same as that of EMIA with the dashed lines excluded.}}
    \label{fig:mia_illustration}
\end{figure*}

\section{Preliminaries}
\label{sec:preliminaries}

In this section, we first introduce the notations and problem setup, and then present a framework typically followed by current MI attacks. We then briefly explain the algorithm of f-LiRA, n-LiRA and EMIA based on the identified framework.

\subsection{Notations and Problem Setup}
We consider a DNN $f: \mathcal{X} \rightarrow [0,1]^m$ that, when given an input $x \in \mathcal{X}$, outputs a vector $f(x)$ of length $m$ containing the probabilities of the class $i$ at the $i$\textsuperscript{th} index $f(x)[i]$. We use $f_t \leftarrow \mathcal{T}(f, D_t)$ to denote that $f$ is trained on some training data $D_t=\{(x^i, y^i)\}_{i=0}^{S-1}$ of size $S$, yielding $f_t$ as the trained DNN. $D_t$ is assumed to be randomly sampled from the real-world data distribution $D_r$, denoted in the future as $D_t \sim D_r \implies \forall (x,y) \in D_t, (x,y) \sim D_r$.

Following~\cite{carlini2022membership}, for any dataset (e.g. CIFAR-10), we take the available training data as $D_r$ and randomly sample $D_t$ from $D_r$. The training phase of the target DNN $f$ is detailed as follows:
\begin{enumerate}
    \item The available training data is assumed to be the real-world data $D_r$ of size $|D_r|$.
    \item Randomly sample training dataset $D_t \sim D_r$ of size $|D_t|=\frac{|D_r|}{2}$.
    \item Output $f_t \leftarrow \mathcal{T}(f, D_t)$.
\end{enumerate}

\subsection{General Membership Inference (MI) Attack Framework}\label{sec:mia}
Given query access to a target DNN $f_t \leftarrow \mathcal{T}(f, D_t)$ trained on the dataset $D_t$, and a set of $k$ subjects $D_e = \left\{ \left( x^i, y^i \right) \right\}_{i=0}^{k-1}$ the membership of which is to be inferred, an MI attack $\mathcal{A}(\cdot)$ aims to find out whether each of the given subjects is a member of $D_t$ or not. An MI attack is based on a \textit{commonly-observed} behavior of DNNs: for any $(x,y) \sim D_r$, the loss of $f_t$ is typically smaller if $(x,y) \in D_t$, than the loss of $f_t$ if $(x,y) \notin D_t$. 

MI attacks assume that the attacker can collect samples from the real-world data $D_r$, from which the training data $D_t$ is sampled. For experimental simulation, MI attacks typically assume that the size of $D_t$ is half that of $D_r$, i.e. $|D_t| = \frac{|D_r|}{2}$). An MI attack generally follows three stages---preparation, indication and decision.

\begin{enumerate}
    \item At the \textit{preparation} stage, given $D_e$, an MI attack prepares the variables $\mathcal{V}$ (e.g., training the models on the attacker dataset $D_a \sim D_r$ sampled from the real-world data) that guide the quantification of membership at the indication stage.
    
    \item At the \textit{indication} stage, $\forall (x,y) \in D_e$ an MI attack uses an indicator function $\mathcal{I}(f_t, (x,y); \mathcal{V})$ guided by $\mathcal{V}$ to compute a value that indicates the membership of the subject $(x,y) \in D_t$.

    The MI attack algorithm $\mathcal{A}$ can thus be characterized by $\mathcal{V}$ and $\mathcal{I}$, such that

    \begin{equation}
        \mathcal{A} (f_t, (x,y)) = \mathcal{I} (f_t, (x,y), \mathcal{V})
    \end{equation}

    Because the objective of an MI attack is to differentiate between $(x,y) \in D_t$ and $(x,y) \notin D_t$, any choice of $\mathcal{V}$ and $\mathcal{I}$ that satisfies the following condition can be used for membership inference.

    \begin{multline}
        \mathcal{V},\mathcal{I},\ \text{such that},\\
        \underset{(x,y) \in D_t}{\mathbb{E}} [ \mathcal{I} (f_t, (x,y), \mathcal{V}) ] > \underset{(x,y) \in D_r \char`\\ D_t}{\mathbb{E}}[ \mathcal{I} (f_t, (x,y), \mathcal{V}) ] 
        \label{eq:general_vi}
    \end{multline}
    
    \item At the \textit{decision} stage, a threshold $\tau$ is decided depending upon the tolerable false positive rate such that $\mathcal{A}(f_t, (x,y)) \geq \tau$, means $(x,y) \in D_t$, and $\mathcal{A}(f_t, (x,y)) < \tau$ means $(x,y) \notin D_t$. Formally,
    \begin{equation}
        b = 
        \begin{cases}
            0, & \mathcal{A}(f_t, (x,y)) < \tau \\
            1, & \mathcal{A}(f_t, (x,y)) \geq \tau
        \end{cases}
        \label{eq:mia_decision}
    \end{equation}
\end{enumerate}

A number of algorithms $\mathcal{A}(\cdot)$ have been proposed to effectively infer the membership of a data sample yielding several MI attacks with diverse threat settings~\cite{shokri2017membership, carlini2022membership, ye2022enhanced, nguyen2023active, jagielski2023students}. In this work, we consider two of the \textit{strongest} and the \textit{most recent} MI attacks: the Likelihood Ratio Attack (LiRA) proposed by Carlini et al.~\cite{carlini2022membership}, and an Enhanced Membership Inference Attack (EMIA) proposed by Ye et al.~\cite{ye2022enhanced}.

\begin{algorithm}[t]
    \footnotesize
    \caption{Offline LiRA (f-LiRA) Algorithm (\cite{carlini2022membership})}
    \label{alg:f-lira}
    \begin{algorithmic}[1]
    
    \Input
    \Statex $f_t \gets$ victim model trained on $D_t$
    \Statex $D_r \gets$ the real data distribution
    \Statex $D_e \gets \left\{ \left( x^i, y^i \sim D_r \right) \right\}_{i=0}^{k+1}$ a set of target samples 
    
    \Output
    \Statex $\mathcal{I} \gets$ membership inference indicator of each target in $D_e$

    \Statex
    \Procedure{$\phi$}{$p$}
        \State \textbf{return} $\log\frac{p}{1-p}$
    \EndProcedure

    \Statex \textcolor{gray}{// preparation stage}
    \For{$j=0..N-1$}
        \State $D^j_a \sim D_r$, such that $\forall (x_a,y_a) \in D^j_a, (x_a,y_a) \notin D_e$
        \State $f^j_n \leftarrow \mathcal{T} \left( f, D^j_a \right)$
    \EndFor
    \State $\mu^i_n \gets \underset{j}{\mathbb{E}} \left[ \phi (x^i,y^i|f_n^j) \right]$, $\sigma^i_n \gets \mathbb{S} \left[ \phi (x^i,y^i|f_n^j) \right]$
    \State $\mathcal{V} \gets \{\mu^i_n, \sigma^i_n\}$

    \Statex \textcolor{gray}{// indication stage}
    \State $\mathcal{I} \gets \{\}$
    \For{$(x^i,y^i)$ \textbf{in} $D_e$}
        \State $LR_f (f_t, (x^i,y^i), \mathcal{V}) \gets 1 - p(\ \phi(x^i,y^i|f_t) \ | \ \mathcal{N}(\mu^i_n, \sigma^i_n) \ )$
        \State $\mathcal{I} \gets \mathcal{I} \cup LR_f (f_t, (x^i,y^i))$
    \EndFor

    \State \textbf{return} $\mathcal{I}$
    \end{algorithmic}
\end{algorithm}

\begin{algorithm}[t]
    \footnotesize
    \caption{Online LiRA (n-LiRA) Algorithm (\cite{carlini2022membership}). Blue highlighted text indicates modifications over f-LiRA.}
    \label{alg:n-lira}
    \begin{algorithmic}[1]
    
    \Input
    \Statex $f_t \gets$ victim model trained on $D_t$
    \Statex $D_r \gets$ the real data distribution
    \Statex $D_e \gets \left\{ \left( x^i, y^i \sim D_r \right) \right\}_{i=0}^{k+1}$ a set of target samples 
    
    \Output
    \Statex $\mathcal{I} \gets$ membership inference indicator of each target in $D_e$

    \Statex \textcolor{gray}{// preparation stage}
    \State $\phi \gets$ Algorithm~\ref{alg:f-lira}
    \For{$(x^i,y^i) \in D_e$}
        \For{$j=0..N-1$}
            \State $D^j_a \sim D_r$, such that $\forall (x_a,y_a) \in D^j_a, (x_a,y_a) \notin D_e$
            \State $f^j_n \leftarrow \mathcal{T} \left( f, D^j_a \right)$, \textcolor{blue}{$f^{i,j}_m \leftarrow \mathcal{T} \left( f, D^j_a \cup (x^i,y^i) \right)$}
        \EndFor
        \State $\mu^i_n \gets \underset{j}{\mathbb{E}} \left[ \phi (x^i,y^i|f_n^j) \right], \sigma^i_n \gets \mathbb{S} \left[ \phi (x^i,y^i|f_n^j) \right]$
        \State \textcolor{blue}{$\mu^i_m \gets \underset{j}{\mathbb{E}} \left[ \phi (x^i,y^i|f_m^{i,j}) \right], \sigma^i_m \gets \mathbb{S} \left[ \phi (x^i,y^i|f_m^{i,j}) \right]$}
        \State $\mathcal{V} \gets \mathcal{V} \cup \{\mu^i_n, \sigma^i_n, \textcolor{blue}{\mu^i_m, \sigma^i_m}\}$
    \EndFor
    
    \Statex \textcolor{gray}{// indication stage}
    \State $\mathcal{I} \gets \{\}$
    \For{$(x^i,y^i)$ \textbf{in} $D_e$}
        \State \textcolor{blue}{$LR_n (f_t, (x^i,y^i), \mathcal{V}) \gets \cfrac{p(\ \phi(x^i,y^i|f_t)\ |\ \mathcal{N}(\mu^i_m, \sigma^i_m) \ )}{p(\ \phi( x^i,y^i|f_t ) \ | \ \mathcal{N}(\mu^i_n, \sigma^i_n) \ )}$}
        \State $\mathcal{I} \gets \mathcal{I} \cup LR_n (f_t, (x^i,y^i), \mathcal{V})$
    \EndFor

    \State \textbf{return} $\mathcal{I}$
    \end{algorithmic}
\end{algorithm}

\vspace{2mm}
\heading{Likelihood Ratio Attack (LiRA):}
LiRA may work in two different modes: offline LiRA and online LiRA.

\headingTwo{Offline LiRA (f-LiRA):} Given $(x^i,y^i) \in D_e$, f-LiRA algorithm is shown in Algorithm~\ref{alg:f-lira}, and summarized below:

\begin{enumerate}
    \item \textit{Preparation (Steps 4 to 9):} 
    \begin{equation}
        \forall j \in [0..N],\ f^j_n \gets \mathcal{T}(D^j_a \sim D_r \char`\\ D_e)
    \end{equation}

    \begin{multline}
        \mathcal{V}:\ 
        \mu^i_n = \underset{j}{\mathbb{E}} \left[ \phi \left( x^i,y^i|f^j_n \right) \right],\ \sigma^i_n = \mathbb{S} \left[ \phi \left( x^i,y^i|f^j_n \right) \right]
    \end{multline}

    where $\underset{j}{\mathbb{E}}$ is the expectation over $j$, $\mathbb{S}$ is the standard deviation and $\phi(x^i,y^i|f)$ is defined as follows,
        \begin{equation}
            \phi \left( x^i,y^i|f \right) = \log \frac{f(x^i)[y^i]}{1-f(x^i)[y^i]}
            \label{eq:phi}
        \end{equation}

    \item \textit{Indication (Steps 10 to 15):} $\forall (x^i,y^i) \in D_e,$
        \begin{multline}
            \mathcal{I}: LR_f (f_t, (x^i,y^i), \mathcal{V}) = \\
            1 - p(\ \phi(x^i,y^i|f_t) \ | \ \mathcal{N}(\mu^i_n, \sigma^i_n) \ )
            \label{eq:f-lira_indicator}
        \end{multline}

    where $LR_f$ denotes the offline likelihood ratio and $p$ is the probability.

    \item \textit{Decision Stage:} The threshold $\tau$ is computed based on the tolerable false positives.
\end{enumerate}

\headingTwo{Online LiRA (n-LiRA):} Online LiRA (n-LiRA) customizes f-LiRA to each subject in $D_e$, and is notably more effective than f-LiRA (Algorithm~\ref{alg:n-lira}).

\begin{enumerate}
    \item \textit{Preparation:}
        \begin{subequations}
        \begin{eqnarray}
            \forall j \in [0..N],\ D^j_a \sim D_r \char`\\ D_e, \nonumber \\
            f^j_n \gets \mathcal{T}(D^j_a)\\
            \forall (x^i,y^i) \in D_e,\ f^j_m \gets \mathcal{T} (D^j_a \cup (x^i,y^i))
        \end{eqnarray}
        \end{subequations}
    
        \begin{multline}
            \mathcal{V}:\ 
            \mu^i_n \gets \underset{j}{\mathbb{E}} \left[ \phi (x^i,y^i|f_n^j) \right], \sigma^i_n \gets \mathbb{S} \left[ \phi (x^i,y^i|f_n^j) \right] \\
            \mu^i_m \gets \underset{j}{\mathbb{E}} \left[ \phi (x^i,y^i|f_m^{i,j}) \right], \sigma^i_m \gets \mathbb{S} \left[ \phi (x^i,y^i|f_n^{i,j}) \right]
            \label{eq:nlira_preparation}
        \end{multline}
    
    \item \textit{Indication:} $\forall (x^i,y^i) \in D_e,$
            \begin{multline}
                \mathcal{I}: LR_n (f_t, (x^i,y^i), \mathcal{V}) = \\
                \cfrac{p(\ \phi(x^i,y^i|f_t)\ |\ \mathcal{N}(\mu^i_m, \sigma^i_m) \ )}{p(\ \phi(x^i,y^i|f_t) \ | \ \mathcal{N}(\mu^i_n, \sigma^i_n) \ )}
                \label{eq:n-lira_indicator}
            \end{multline}

        where $LR_n$ denotes the online likelihood ratio.
    
    \item \textit{Decision:} Finally, the threshold $\tau$ is computed based on the tolerable false positives.
\end{enumerate}

\begin{algorithm}[t]
    \footnotesize
    \caption{EMIA Algorithm (\cite{ye2022enhanced}). Blue highlighted text indicates modifications over f-LiRA.}
    \label{alg:emia}
    \begin{algorithmic}[1]
    
    \Input
    \Statex $f_t \gets$ victim model trained on $D_t$
    \Statex $D_r \gets$ the real data distribution
    \Statex $D_e \gets \left\{ \left( x^i, y^i \sim D_r \right) \right\}_{i=0}^{k+1}$ a set of target samples 
    
    \Output
    \Statex $\mathcal{I} \gets$ membership inference indicator of each target in $D_e$

    \Statex
    \Statex \textcolor{gray}{// preparation stage}
    \State $\phi \gets$ Algorithm~\ref{alg:f-lira}
    \For{$j=0..N-1$}
        \State \textcolor{blue}{$D^{(j)}_a \sim D_r$, s.t. $\forall (x_a,y_a) \in D^{(j)}_a, x_a \neq x$ and $y_a = f_t(x_a)$} \label{alg:step_emiaSampling}
        \State $f^j_n \leftarrow \mathcal{T} \left( f, D^j_a \right)$
    \EndFor
    \State $\mu^i_n \gets \underset{j}{\mathbb{E}} \left[ \phi (x^i,y^i|f_n^j) \right]$, $\sigma^i_n \gets \mathbb{S} \left[ \phi (x^i,y^i|f_n^j) \right]$
    \State $\mathcal{V} \gets \{\mu^i_n, \sigma^i_n\}$

    \Statex \textcolor{gray}{// indication stage}
    \State $\mathcal{I} \gets \{\}$
    \For{$(x^i,y^i)$ \textbf{in} $D_e$}
        \State $LR_f (f_t, (x^i,y^i), \mathcal{V}) \gets 1 - p(\ \phi(x^i,y^i|f_t) \ | \ \mathcal{N}(\mu^i_n, \sigma^i_n) \ )$
        \State $\mathcal{I} \gets \mathcal{I} \cup LR_f (f_t, (x^i,y^i))$
    \EndFor

    \State \textbf{return} $\mathcal{I}$
    \end{algorithmic}
\end{algorithm}

\heading{Enhanced Membership Inference Attack (EMIA):} The working of EMIA is given in Algorithm~\ref{alg:emia}, and summarized below.

\begin{enumerate}
    \item \textit{Preparation (Steps 2 to 7):}
        \begin{multline}
            \forall j \in [0..N],\\
            D^j_a \sim D_r \char`\\ D_e\ \text{s.t.}\ \forall (x_a,y_a) \in D^j_a,\  y_a=f_t(x_a), \\
            f^j_n \gets \mathcal{T}(D^j_a)
        \end{multline}
        \begin{multline}
        \mathcal{V}:\ 
        \mu^i_n = \underset{j}{\mathbb{E}} \left[ \phi \left( x^i,y^i|f^j_n \right) \right],\ \sigma^i_n = \mathbb{S} \left[ \phi \left( x^i,y^i|f^j_n \right) \right]
    \end{multline}
        

    \item The \textit{indication and decision} stages of EMIA are the same as those of f-LiRA.
\end{enumerate}

\section{Proposed Membership Inference Attacks}\label{sec:methodology}

In this section, we present our algorithm to prepare the variables for membership inference with two novel improvements---firstly, our attack utilizes both the membership and the non-membership information of the subjects at a significantly higher computational efficiency compared to n-LiRA; secondly, our attack optimally leverages the loss landscape around the subjects by computing small magnitude input perturbations, adversarially optimized to maximize the expected loss of non-member shadow DNNs and minimize that of the member shadow DNNs on the subjects. We then present two novel augmented indicators that examine the loss values in the Gaussian neighborhood of the subjects, in addition to the loss over each subject, to infer the membership of the subjects.

\subsection{Adversarial Membership Inference Attack (AMIA)}\label{sec:methodology_amia}
Motivated by eq-\eqref{eq:general_vi}, for each subject $(x^i,y^i) \in D_e$, we compute a perturbation $\Delta x^i$ such that $(x^i + \Delta x^i,y^i)$ is more vulnerable to MI attack as compared to $(x^i, y^i)$. More formally, we aim to achieve the following,

\begin{equation}
\begin{split}
    \mathcal{A}(f_t, (x^i+\Delta x^i,y^i)) \geq \mathcal{A}(f_t, (x^i,y^i)),\ \text{if}\ (x^i,y^i) \in D_t \\
    \mathcal{A}(f_t, (x^i+\Delta x^i,y^i)) \leq \mathcal{A}(f_t, (x^i,y^i)),\ \text{if}\ (x^i,y^i) \notin D_t
\end{split}
    \label{eq:amia_goal}
\end{equation}

However, it is impossible to directly achieve eq-\eqref{eq:amia_goal} because an MI attacker does not have access to $D_t$.
To address this, we leverage the transferability property of adversarial perturbations. It has been shown that the adversarial perturbation to an input computed to fool a deep learning model $\mathcal{F}$ effectively transfers (fools) to other DNNs~\cite{ali2021all}. Additionally, using an ensemble of multiple DNNs as $\mathcal{F}$ significantly improves the transferability of the adversarial perturbations. We leverage the transferability property to transfer the adversarial perturbations computed over member and non-member shadow DNNs to $f_t$ as detailed in Algorithm~\ref{alg:amia}.

\begin{enumerate}
    \item \textit{Preparation Stage:} Given a set of subjects $D_e$ for membership inference, at every iteration $j \in [0..N-1]$, we first randomly sample a dataset $D^j_{a} \sim D_r$ (Step 3) and randomly bisect $D_e$ into two distinct subsets $D^j_{e1}$ and $D^j_{e2}$ (Step 4). We then train a pair of shadow DNNs $f^j_{m1}$ and $f^j_{m2}$ on $D^j_a$ augmented with $D^j_{e1}$ and $D^j_{e2}$ respectively (Step 5). This lets us train the member and the non-member shadow DNNs simultaneously---$\forall (x^i,y^i) \in D_e$, if $(x^i,y^i) \in D^j_{e1}$, $f^j_{m1}$ is the member shadow DNN and $f^j_{m2}$ is the non-member shadow DNN, and vice versa if $(x^i,y^i) \in D^j_{e2}$. The process is repeated for $j \in [0..N-1]$.
    
    To make the attack stronger, $\forall j \in [0..N-1], \forall (x^i,y^i) \in D_e$, we first create lists of member shadow DNNs $f^{i,j}_m$ and non-member shadow DNNs $f^{i,j}_n$ (Steps 10 to 13). We then adversarially learn $\Delta x^i \in [-\epsilon, \epsilon]$, using the iterative Fast Gradient Sign Method (i-FGSM) to minimize the expected loss of $(x^i,y^i)$ on the member shadow DNNs, while maximizing the loss of $(x^i,y^i)$ on non-member shadow DNNs (Step 16).
    Formally,
    
    \begin{multline}
            g^j(\Delta x^i) = \\
            \phi \left( x^i+\Delta x^i, y^i|f^{i,j}_n \right) - \phi\left( x^i+\Delta x^i, y^i|f^{i,j}_m \right) \\
    \end{multline}
    
    \begin{equation}
        \Delta x^i = \min_{\Delta x^i \in [-\epsilon, \epsilon]} \quad \underset{j \in [0..N-1]}{\mathbb{E}} \left[ g^j(\Delta x^i) \right]\\
        \label{eq:amia_loss}
    \end{equation}

    The prepared variables $\mathcal{V}$ are then,
    \begin{multline}
            \mathcal{V}:\ \Delta x^i,\\
            \mu^i_n = \underset{j}{\mathbb{E}} \left[ \phi \left( x^i+\Delta x^i,y^i|f^{i,j}_n \right) \right], \\
            \sigma^i_n = \mathbb{S} \left[ \phi \left( x^i+\Delta x^i,y^i|f^{i,j}_n \right) \right], \\
            \mu^i_m = \underset{j}{\mathbb{E}} \left[ \phi \left( x^i+\Delta x^i,y^i|f^{i,j}_m \right) \right], \\
            \sigma^i_m = \mathbb{S} \left[ \phi \left( x^i+\Delta x^i,y^i|f^{i,j}_m \right) \right]
            \label{eq:amia_preparation}
        \end{multline}

    \item \textit{Indication Stage:} $\forall (x^i,y^i) \in D_e$, and corresponding $i \in [0..|D_e|-1]$ AMIA computes the $\phi$-scaled confidences of non-member and member shadow DNNs on the subject $(x,y)$ and compute $\mathcal{I}$ using the online likelihood ratio $LR_n$ as follows,
    \begin{multline}
        \mathcal{I}: LR_n (f_t, (x^i,y^i), \mathcal{V}) = \\
        \cfrac{p(\ \phi(x^i+\Delta x^i,y^i|f_t)\ |\ \mathcal{N}(\mu^i_m, \sigma^i_m) \ )}{p(\ \phi(x^i+\Delta x^i,y^i|f_t) \ | \ \mathcal{N}(\mu^i_n, \sigma^i_n) \ )}
    \end{multline}

    \item \textit{Decision Stage:} Finally, the threshold $\tau$ is computed based on the tolerable false positives.
\end{enumerate}

\begin{algorithm}[t]
    \footnotesize
    \caption{Proposed Adversarial MIA (AMIA) algorithm. Blue highlighted text indicates modifications over f-LiRA.}
    \label{alg:amia}
    \begin{algorithmic}[1]
    
    \Input
    \Statex $f_t \gets$ victim model trained on $D_t$
    \Statex $D_r \gets$ the real data distribution
    \Statex $D_e \gets \left\{ \left( x^i, y^i \sim D_r \right) \right\}_{i=0}^{k+1}$ a set of target samples 
    
    \Output
    \Statex $\mathcal{I} \gets$ membership inference indicator of each target in $D_e$

    \Statex \textcolor{gray}{// preparation stage}
    \State $\mathcal{V} \gets \{\}$, $\phi \gets$ Algorithm~\ref{alg:f-lira}
    \For{$j=0..N-1$}
        \State $D^j_a \sim D_r$, such that $\forall (x_a,y_a) \in D^j_a, (x_a,y_a) \neq (x,y)$. \label{alg:step_amiaSampling}
        {\color{blue}
        \State $D^j_{e1}$, $D^j_{e2} \gets$ random bisect $(D_e)$
        \State $f^j_{m1} \gets \mathcal{T} \left( f, D^j_a \cup D^j_{e1} \right), f^j_{m2} \gets \mathcal{T} \left( f, D^j_a \cup D^j_{e2} \right)$
        }
    \EndFor
    
    {\color{blue}
    \For{$(x^i,y^i) \in D_e$}
        \For{$j \in [0..N-1]$}
            \State $\Delta x^i \gets 0$
            \If{$(x^i,y^i) \in D^j_{e1}$}
                \State $s_j \gets -1$, $f^{i,j}_m \gets f^j_{m1}$, $f^{i,j}_n \gets f^j_{m2}$
            \ElsIf{$(x^i,y^i) \in D^j_{e2}$}
                \State $s_j \gets 1$, $f^{i,j}_m \gets f^j_{m2}$, $f^{i,j}_n \gets f^j_{m1}$
            \EndIf
        \EndFor
    
        \Statex
        \State $\Delta x^i \gets \underset{\Delta x^i}{\min}\ \underset{j}{\mathbb{E}} \left[ \phi \left( x^i+\Delta x^i, y^i|f^{i,j}_n \right) - \phi\left( x^i+\Delta x^i, y^i|f^{i,j}_m \right) \right]$
        \State $\mu^i_n \gets \underset{j}{\mathbb{E}} \left[ \phi (x^i,y^i|f_n^{i,j}) \right], \sigma^i_n \gets \mathbb{S} \left[ \phi (x^i,y^i|f_n^{i,j}) \right]$
        \State $\mu^i_m \gets \underset{j}{\mathbb{E}} \left[ \phi (x^i,y^i|f_m^{i,j}) \right], \sigma^i_m \gets \mathbb{S} \left[ \phi (x^i,y^i|f_m^{i,j}) \right]$
        \State $\mathcal{V} \cup \{\Delta x^i, \mu^i_n, \mu^i_m, \sigma^i_n, \sigma^i_m\}$
    \EndFor
    }

    {\color{blue}
    \Statex \textcolor{gray}{// indication stage}
    \State $\mathcal{I} \gets \{\}$
    \For{$(x^i,y^i) \in D_e$}
        \State $LR_n (f_t, (x^i,y^i), \mathcal{V}) \gets \cfrac{p(\ \phi(x^i+\Delta x^i,y^i|f_t) \ | \ \mathcal{N}(\mu^i_m, \sigma^i_m) \ )}{p(\ \phi(x^i + \Delta x^i,y^i|f_t) \ | \ \mathcal{N}(\mu^i_n, \sigma^i_n) \ )}$
        \State \textcolor{black}{$\mathcal{I} \gets \mathcal{I} \cup LR_n (f_t, (x^i,y^i), \mathcal{V})$}
    \EndFor
    }
    \State \textbf{return} $\mathcal{I}$
    \end{algorithmic}
\end{algorithm}

\vspace{1mm}
\noindent \textbf{Enhanced Adversarial Membership Inference Attack. \space}
It may be possible to further increase the effectiveness of AMIA by reducing the data uncertainty inspired by EMIA. To achieve this, we propose the enhanced adversarial membership inference attack (E-AMIA). E-AMIA follows the same algorithm as AMIA, except that it replaces the data sampling step~\ref{alg:step_amiaSampling} of Algorithm~\ref{alg:amia} with the data sampling step~\ref{alg:step_emiaSampling} of Algorithm~\ref{alg:emia}.

\subsection{Augmented Membership Indicators}
Both LiRA and EMIA use \textit{vanilla indicators} that examine the loss of the target DNN precisely upon the subject $(x^i,y^i)$. However, loss values in the local neighborhood (e.g., the Gaussian neighborhood) of the subject might be helpful for membership inference. In order to utilize the Gaussian neighborhood information of a subject, we create a set of augmented inputs $X^i$ by concatenating the original input $x$ with its Gaussian noise augmented versions that are computed by adding to $x^i$, $p-1$ random Gaussian noises $\bigcup_{l=1}^{p-1} n^l \sim \mathcal{N}(0, \sigma_n)$ with mean $0$ and standard deviation $\sigma_n$.

\begin{equation}
    X^i = \bigcup_{l=0}^{p-1} \left( x^i + n^l \sim \mathcal{N}(0, \sigma_n) \right)
\end{equation}

where $n^{(0)} = 0$. This lets us analyze the loss of $f_t$ at and around $(x^i,y^i)$ allowing us to model local trends which are helpful in membership inference (See Fig.).

\subsubsection{Likelihood Ratio with Perturbation:}
For any $(x^i,y^i) \in D_e$, follow the following steps:

\begin{enumerate}
    \item $\forall l \in [0..p-1]$, compute
        \begin{equation}\begin{split}
            F_n^l = \left\{ \forall j \in [0..N-1], \phi \left( x^i + n^l,y|f^j_n \right) \right\} \\
            F_m^l = \left\{ \forall j \in [0..N-1], \phi \left( x^i + n^l,y|f^j_m \right) \right\}
        \end{split}\end{equation}
    
    \item $\forall l \in [0..p-1]$, compute
        \begin{equation}\begin{split}
            \mu_n^l = \operatorname{mean} \left( F_n^l \right), \sigma_n^l = \operatorname{std} \left( F_n^l \right) \\
            \mu_m^l = \operatorname{mean} \left( F_m^l \right), \sigma_m^l = \operatorname{std} \left( F_m^l \right)
        \end{split}\end{equation}

    \item Finally, compute $LR_p (f_t, (x^i,y^i), \mathcal{V})$ the \textit{likelihood ratio of $(x^i,y^i)$ to be a member of $D_t$ with perturbation} defined as the expected likelihood ratio of $(x^i+n^l,y^i)$ over $l$ for $f_t$.
    \begin{multline}
        LR_p (f_t, (x,y), \mathcal{V}) = \\
        \underset{l \in [0..p-1]}{\mathbb{E}} \left[ LR_n \left( f_t, \left( x^i+n^l, y^i \right), F_n^l, F_m^l \right) \right]
        \label{eq:lr_p}
    \end{multline}
\end{enumerate}

\subsubsection{Likelihood Ratio with the Optimized Perturbation:}
For any $(x^i,y^i) \in D_e$, follow the following steps:

\begin{enumerate}
    \item Follow steps 1 and 2 of the likelihood ratio with perturbation.

    \item Set $O = \{\}$,
    \begin{equation}
        \text{Repeat $z$ times:}\quad O := O \cup \underset{l \in [0..p-1], l \notin O}{\operatorname{argmax}} \left( F_m^l - F_n^l \right)
    \end{equation}

    $O$ is a set of $z$ integers denoting the indices $l \in [0..p-1]$ of the noise that produce the maximum expected difference between the $\phi$-scaled confidences of the member and the non-member shadow DNNs.

    \item Finally, compute $LR_o (f_t, (x^i,y^i))$ the \textit{likelihood ratio of $(x^i,y^i)$ to be a member of $D_t$ with optimal perturbation} defined as the expected likelihood ratio of $(x+n^l,y)$ over $l \in O$ for $f_t$.
    \begin{equation}
    \resizebox{.9\hsize}{!}{$
        LR_o (f_t, (x,y)) = \underset{l \in O}{\mathbb{E}} \left[ LR_n \left( f_t, \left( x^i+n^l, y^i \right), F_n^l, F_m^l \right) \right]
    $}
        \label{eq:lr_o}
    \end{equation}
    
    Note that if $z = p$, $LR_o(\cdot) = LR_n(\cdot)$.
\end{enumerate}

\section{Performance Evaluation}\label{sec:results}

\subsection{Evaluation Methodology}
Our evaluation methodology is similar to that used by the previous works~\cite{carlini2022membership, ye2022enhanced} as summarized below.

\vspace{2mm}
\subsubsection{Datasets}
We evaluate MI attacks on three commonly used machine learning datasets---CIFAR-10, MNIST and Fashion-MNIST. All our datasets are divided into two sets: (1) the original training set of the data is assumed to be our real-world data $D_r$, and (2) the original test set is denoted as $D_s$. 
The training data $D_t \sim D_r$ for the target DNN is sampled from $D_r$. For coherence with Carlini et al.~\cite{carlini2022membership}, we use $|D_t| = |D_r|/2$, where $|\cdot|$ denotes the size of the dataset. The size of the attacker data $D_a \sim D_r$ is $|D_a|=|D_t|/2$~\cite{carlini2022membership}. We construct the evaluation dataset $D_e$ of size $|D_e| = k$, where each instance is sampled uniformly either from $D_t$ or $D_r \char`\\ D_t$. We use $k=200$ in our experiments.

\vspace{2mm}
\subsubsection{Model architecture}
Following Ye et al.~\cite{ye2022enhanced}, we chose a deep Convolutional Neural Network (CNN)-based model as our target DNN. More specifically, we use the following CNN: \textit{Input() - \{Conv2D() - ReLU() - BatchNorm()\}$\times 3$ - MaxPool() - Dropout(.) - Flatten() - Dense($\# classes$) - softmax()}. Unless stated otherwise, we always train the CNN using $L_2$ regularization because it makes the CNNs notably more robust to MI attacks~\cite{choquette2021label}. We train $N=50$ non-member shadow DNNs for f-LiRA and EMIA, and equal number of non-member and member shadow DNNs for AMIA and E-AMIA for each experiment. 
We do not implement n-LiRA due to its high computational overhead---n-LiRA requires training 10000 shadow DNNs for a single experiment on each dataset in our experimental setup.


\begin{figure}
    \centering
    \includegraphics[width=0.6\linewidth]{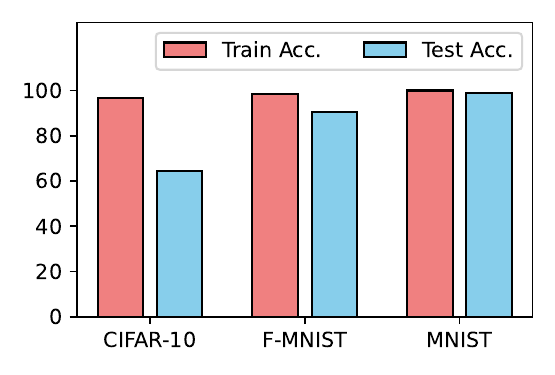}
    \caption{Comparing the train and test accuracies of the CNN on the three datasets considered in this paper.}
    \label{fig:accuracies}
\end{figure}

\vspace{2mm}
\subsubsection{Evaluation Metrics}
\label{sec:setup_metrics}
Here we first define the TPR-FPR curve typically used by prior works~\cite{carlini2022membership, ye2022enhanced, liu2022membership}, and then present the running TPR average (RTA), a novel metric to test the performance of MI attacks. Although AMIA and E-AMIA generally perform better than f-LiRA and EMIA on both evaluation metrics, we believe that RTA better distinguishes different attacks as compared to the TPR-FPR curve.

\subsubsection{TPR-FPR curve}
Following previous works we evaluate the effectiveness of an MI attack by computing its TPR at a given FPR in log scale (emphasizing low FPR region)~\cite{carlini2022membership}. Given $\tau$, TPR is the ratio of the number of samples correctly marked as members by $\mathcal{A}$, to the total number of member samples. Similarly, the FPR is the total number of samples incorrectly marked by $\mathcal{A}$ as members to the total number of non-member samples. Both TPR and FPR are define below:

\begin{equation}
    \text{TPR} (\tau) = \frac{|\forall (x,y) \in D_t \cap D_e,\ s.t. \ \mathcal{A}(f_t, (x,y)) > \tau|} {|\forall (x,y) \in D_t \cap D_e|}
    \label{eq:tpr}
\end{equation}
\begin{equation}
    \text{FPR} (\tau) = \frac{|\forall (x,y) \in \{D_r \char`\\ D_t\} \cap D_e,\ s.t. \ \mathcal{A}(f_t, (x,y)) > \tau|} {|\forall (x,y) \in D' \cap D_e|}
    \label{eq:fpr}
\end{equation}

where, $|\cdot|$ denotes the total number of samples satisfying the condition.

\subsubsection{Running TPR Average (RTA)}
We propose a new metric called the running TPR average (RTA) defined at a given value $t \in [0,1]$ as the average value of TPR for FPR less than $t$. Formally,

\begin{equation}
    \text{RTA} (t) = \underset{\tau\ \text{s.t.}\ \text{FPR}(\tau) \leq t}{\mathbb{E}} [ \text{TPR}(\tau) ]
\end{equation}

Stated simply, RTA$(t)$ defines the success rate of membership detection rate with more than $(1-t) \times 100\%$ confidence. We believe that RTA better quantifies the member inference efficacy of an MI attack in the low FPR region. Nevertheless, we show that our proposed attacks---AMIA and E-AMIA---outperform the state-of-the-art on both of the evaluation metrics.







\subsection{Accuracy of the Models}
On CIFAR-10 dataset, the CNN achieves an accuracy of 96.7\% accuracy on the training dataset $D_t$ with 64.6\% accuracy on the test set $D_s$ as shown in Fig.~\ref{fig:accuracies}, which is the same as that achieved by Carlini et al.~\cite{carlini2022membership} using wide ResNet. For F-MNIST and MNIST, Fig.~\ref{fig:accuracies} show that the CNN respectively achieves 98.2\% and 99.9\% accuracy on $D_t$, and 90.6\% and 98.6\% accuracy on $D_s$. Like Carlini et al.~\cite{carlini2022membership}, we achieve a significantly smaller accuracy on the test set as compared to the state-of-the-art for CIFAR-10 and F-MNIST. This is because the experimental setup of LiRA (and EMIA) further divides the original training set $D_r$ into two equal halves, and only one of these halves $D_t$ is used to train the CNN.

\begin{figure*}
    \centering
    \begin{subfigure}{0.32\linewidth}
        \centering
        \includegraphics[page=1, width=\linewidth]{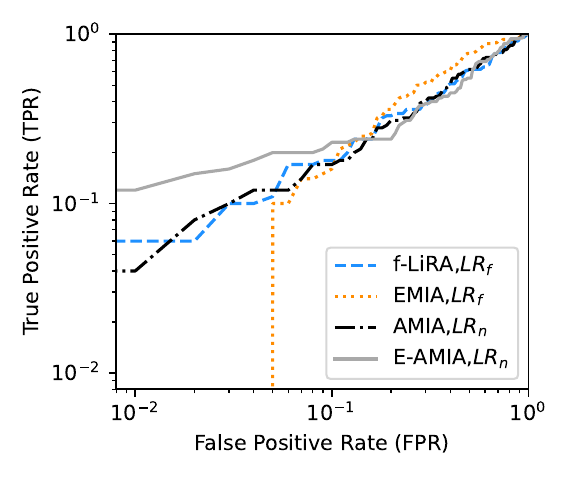}
        \caption{CIFAR-10}
    \end{subfigure}
    \begin{subfigure}{0.32\linewidth}
        \centering
        \includegraphics[page=2, width=\linewidth]{Figures/standard0.02,0.02.pdf}
        \caption{Fashion-MNIST}
    \end{subfigure}
    \begin{subfigure}{0.32\linewidth}
        \centering
        \includegraphics[page=3, width=\linewidth]{Figures/standard0.02,0.02.pdf}
        \caption{MNIST}
    \end{subfigure}
    
    \caption{Comparing TPR of the state-of-the-art MI attacks (f-LiRA,$LR_f$ and EMIA,$LR_f$) with our proposed attacks (AMIA,$LR_n$ and E-AMIA,$LR_n$) for the entire range of possible FPR in log scale to emphasize low FPR region~\cite{carlini2022membership, ye2022enhanced}. \take{Both AMIA,$LR_n$ and E-AMIA,$LR_n$ perform notably better than f-LiRA,$LR_f$ and EMIA,$LR_f$ in low FPR region.} \settings{$
    \epsilon=0.02$, $N=50$}. }
    \label{fig:tpr_fpr}
\end{figure*}
\begin{figure*}
    \centering
    \begin{subfigure}{0.32\linewidth}
        \centering
        \includegraphics[page=1, width=\linewidth]{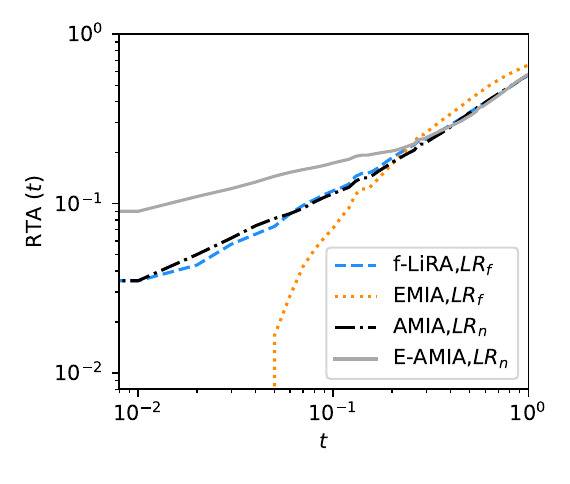}
        \caption{CIFAR-10}
    \end{subfigure}
    \begin{subfigure}{0.32\linewidth}
        \centering
        \includegraphics[page=2, width=\linewidth]{Figures/running_average0.02,0.02.pdf}
        \caption{Fashion-MNIST}
    \end{subfigure}
    \begin{subfigure}{0.32\linewidth}
        \centering
        \includegraphics[page=3, width=\linewidth]{Figures/running_average0.02,0.02.pdf}
        \caption{MNIST}
    \end{subfigure}
    
    \caption{Comparing the running TPR average (RTA) of the state-of-the-art MI attacks (f-LiRA,$LR_f$ and EMIA,$LR_f$) with our proposed attacks (AMIA,$LR_n$ and E-AMIA,$LR_n$) as $t$ increases logarithmically to emphasize low FPR region~\cite{carlini2022membership, ye2022enhanced}. \take{Both AMIA,$LR_n$ and E-AMIA,$LR_n$ perform notably better than f-LiRA,$LR_f$ and EMIA,$LR_f$ in low FPR region. RTA more effectively distinguishes different attacks in the low FPR region.} \settings{$\epsilon=0.02$, $N=50$}. }
    \label{fig:ra_fpr}
\end{figure*}

\subsection{Evaluating MI Attacks}\label{sec:results_evaluation}
This section compares the state-of-the-art MI attacks---f-LiRA and EMIA---with the newly proposed AMIA and E-AMIA. Throughout the section (and the following sections), we denote an MI attack using the $\mathcal{V}, \mathcal{I}$ format. For example, f-LiRA,$LR_f$ denotes that f-LiRA algorithm is used to prepare the variables and $LR_f$ is used as the indicator function.

\subsubsection{Comparing MI attacks}
Comparison of f-LiRA,$LR_f$~\cite{carlini2022membership}, EMIA,$LR_f$~\cite{ye2022enhanced}, AMIA,$LR_n$ (ours) and E-AMIA,$LR_n$ (ours) respectively for the CIFAR-10, Fashion-MNIST and MNIST datasets is provided in Fig.~\ref{fig:tpr_fpr}(a)-(c) based on TPR-FPR metric, and in Fig.~\ref{fig:ra_fpr}(a)-(c) based on RTA metric. We note that f-LiRA,$LR_f$ is more effective than EMIA,$LR_f$ in the low FPR region---Ye et al.~\cite{ye2022enhanced} also observe that Attack R (f-LiRA,$LR_f$ in this paper) performs better than Attack D (EMIA,$LR_f$ in this paper) in the low FPR region on CIFAR-10 dataset. In the high FPR region, EMIA,$LR_f$ typically outperforms including the proposed AMIA and E-AMIA---in their work, Ye et al.~\cite{ye2022enhanced} also observe that EMIA,$LR_f$ is more effective in the high FPR region on CIFAR-10 dataset. We also note that, although our general observations and conclusions from RTA and TPR-FPR curve are consistent with each other, RTA is able to better distinguish the effectiveness of different attacks and appears more stable as compared to the TPR-FPR curve.

As we are mostly interested in membership detection with high confidence (low FPR region)~\cite{carlini2022membership}, our results show that AMIA,$LR_n$ and E-AMIA,$LR_n$ consistently outperform both f-LiRA,$LR_f$ and EMIA,$LR_f$ for low FPRs (e.g., $\leq$ 3\%) with a significant margin for all the three datasets. The only exception is CIFAR-10 at FPR 1\%, where f-LiRA,$LR_f$ performs slightly better. For example, on the Fashion-MNIST dataset, f-LiRA,$LR_f$ and EMIA,$LR_f$ exhibit 0\% TPR, while AMIA,$LR_n$ and E-AMIA,$LR_n$ respectively achieve 6\% and 8\% TPR at 1\% FPR. This is concerning as it implies that around 8\% of the training dataset members can be identified with 99\% confidence (1\% FPR). We attribute this increased effectiveness to two reasons. Firstly, AMIA,$LR_n$ is guided by both the membership and the non-membership properties of the subject $(x^i,y^i) \in D_e$ in contrast to EMIA,$LR_f$, which is only informed by the non-membership properties. Secondly, AMIA,$LR_f$ leverages small adversarial perturbations (or adversarial features) that have been carefully optimized to be indicative of the membership of $(x,y)$.

\begin{table*}[t]
\centering
\caption{Comparing RTA$(\tau)$ of f-LiRA, EMIA, AMIA and E-AMIA for multiple indicator functions. $\ast$ denotes the newly proposed attacks and indicator functions in this paper. Red highlighted values indicate instances where our proposed attack or indicator functions perform worse than either LiRA or EMIA.}
\resizebox{0.9\linewidth}{!}{%
\begin{tabular}{|c|c|cc|cc|ccc|ccc|}
\hline
\multirow{2}{*}{\textbf{Dataset}} & \multirow{2}{*}{\textbf{Metric}} &
\multicolumn{2}{c|}{\textbf{f-LiRA}} &
\multicolumn{2}{c|}{\textbf{EMIA}} &
\multicolumn{3}{c|}{\textbf{AMIA $\ast$}} &
\multicolumn{3}{c|}{\textbf{E-AMIA $\ast$}} \\
& & $LR_f$ & $LR_p$\textbf{$\ast$} & $LR_f$ & $LR_p$\textbf{$\ast$} & $LR_f$ & $LR_p$\textbf{$\ast$} & $LR_o$\textbf{$\ast$} & $LR_f$ & $LR_p$\textbf{$\ast$} & $LR_o$\textbf{$\ast$} \\
\hline
\multirow{5}{*}{\textbf{CIFAR-10}} 
& RTA(0) & 0.01 & 0.01 & 0.0 & 0.02 & 0.03 & 0.03 & 0.02 & \textbf{0.06} & \underline{\textbf{0.07}} & 0.04\\
& RTA(0.01) & 0.035 & \textcolor{red}{0.03} & 0.0 & 0.025 & 0.035 & 0.04 & \textcolor{red}{0.025} & \underline{\textbf{0.09}} & \underline{\textbf{0.09}} & \textbf{0.075}\\
& RTA(0.03) & 0.0575 & \textcolor{red}{0.0425} & 0.0 & 0.0325 & 0.0625 & 0.07 & \textcolor{red}{0.0575} & 0.1225 & \textbf{0.13} & \underline{\textbf{0.1325}}\\
& RTA(0.05) & 0.07333 & \textcolor{red}{0.065} & 0.01667 & 0.05333 & 0.08167 & 0.08667 & 0.07667 & 0.145 & \textbf{0.15} & \underline{\textbf{0.155}}\\
& RTA(1) & 0.5708 & \textcolor{red}{0.5546} & \underline{\textbf{0.6554}} & \textcolor{red}{\textbf{0.6483}} & \textcolor{red}{0.5749} & \textcolor{red}{0.572} & \textcolor{red}{0.5703} & \textcolor{red}{0.5767} & \textcolor{red}{0.5845} & \textcolor{red}{0.586}\\

\hline
\multirow{5}{*}{\textbf{Fashion-MNIST}} 
& RTA(0) & 0.0 & 0.0 & 0.0 & \textbf{0.01} & \textbf{0.01} & \textbf{0.01} & 0.0 & \textbf{0.01} & \underline{\textbf{0.02}} & \underline{\textbf{0.02}}\\
& RTA(0.01) & 0.0 & 0.0 & 0.0 & \textbf{0.045} & 0.035 & 0.04 & 0.035 & \textbf{0.045} & \underline{\textbf{0.055}} & 0.04\\
& RTA(0.03) & 0.0025 & 0.0075 & 0.0 & 0.0625 & 0.0525 & 0.0575 & 0.055 & 0.0625 & \underline{\textbf{0.0825}} & \textbf{0.075}\\
& RTA(0.05) & 0.01333 & 0.015 & 0.02 & 0.07167 & 0.06167 & 0.065 & 0.06333 & 0.08167 & \underline{\textbf{0.09167}} & \textbf{0.09}\\
& RTA(1) & 0.4852 & 0.5057 & \underline{\textbf{0.5262}} & \textcolor{red}{\textbf{0.5259}} & \textcolor{red}{0.4857} & \textcolor{red}{0.486} & \textcolor{red}{0.4927} & \textcolor{red}{0.5225} & \textcolor{red}{0.5192} & \textcolor{red}{0.5154}\\

\hline
\multirow{5}{*}{\textbf{MNIST}} 
& RTA(0) & 0.01 & 0.02 & 0.0 & 0.0 & \underline{\textbf{0.06}} & \underline{\textbf{0.06}} & \underline{\textbf{0.06}} & \textbf{0.04} & 0.03 & 0.03\\
& RTA(0.01) & 0.015 & 0.025 & 0.0 & 0.005 & \underline{\textbf{0.06}} & \underline{\textbf{0.06}} & \underline{\textbf{0.06}} & \textbf{0.04} & 0.03 & 0.03\\
& RTA(0.03) & 0.025 & 0.0275 & 0.0 & 0.0225 & \underline{\textbf{0.07}} & \underline{\textbf{0.07}} & \underline{\textbf{0.07}} & \textbf{0.0475} & 0.0425 & 0.045\\
& RTA(0.05) & 0.03167 & \textcolor{red}{0.03} & 0.02 & 0.04167 & \underline{\textbf{0.07333}} & \underline{\textbf{0.07333}} & \underline{\textbf{0.07333}} & 0.06333 & \textbf{0.06333} & \textbf{0.06333}\\
& RTA(1) & 0.4877 & \textcolor{red}{0.4636} & 0.5353 & \textcolor{red}{0.5192} & 0.5368 & 0.5368 & 0.5377 & 0.5841 & \textbf{0.5857} & \underline{\textbf{0.5876}}\\
\hline
\end{tabular}%
}
\label{tab:indicators}
\end{table*}
\subsubsection{Comparing indicator functions}
We compare the performance of different indicator functions $\mathcal{I}$ with f-LiRA, EMIA, AMIA and E-AMIA on the three datasets considered in this paper. More specifically, for f-LiRA and EMIA, we compare two indicators $LR_f$ and $LR_p$, and for AMIA, we compare three indicators $LR_n$, $LR_p$ and $LR_o$, because $LR_o$ is only compatible with the algorithms that leverage the membership information of $(x,y)$ (See eq-\eqref{eq:lr_o}) in Table~\ref{tab:indicators}. 

Carlini et al.~\cite{carlini2022membership} observe that Gaussian augmented indicator proposed by Jayaraman et al.~\cite{jayaraman2021revisiting} does not improve the MI attack performance. On the contrary, our proposed augmented indicators $LR_p$ and $LR_o$ typically outperform vanilla indicators $LR_f$ and $LR_n$ for all the baseline attacks, specifically on Fashion-MNIST and MNIST datasets. This is because our novel augmented indicators compute the membership likelihood of each noisy sample independently, and then ensemble the results as defined in eq-\eqref{eq:lr_p} and \eqref{eq:lr_o}. These observations are also consistent with our initial hypothesis that analyzing local loss at and around $(x,y)$ will improve the effectiveness of MI attacks.

Table~\ref{tab:indicators} also validates our previous finding that AMIA outperforms both f-LiRA and EMIA in the low FPR region, while EMIA is most effective for the high FPR region (see AUC(1) in the table).

\subsection{Analyzing Hyperparameters}\label{sec:results_hyperparameters}

\begin{figure*}
    \centering
    \includegraphics[page=1, width=0.85\linewidth]{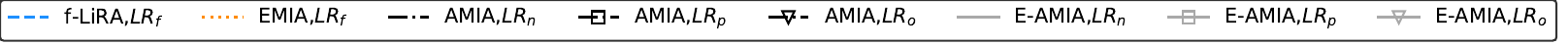}
    \begin{subfigure}{0.32\linewidth}
        \centering
        \includegraphics[page=1, width=\linewidth]{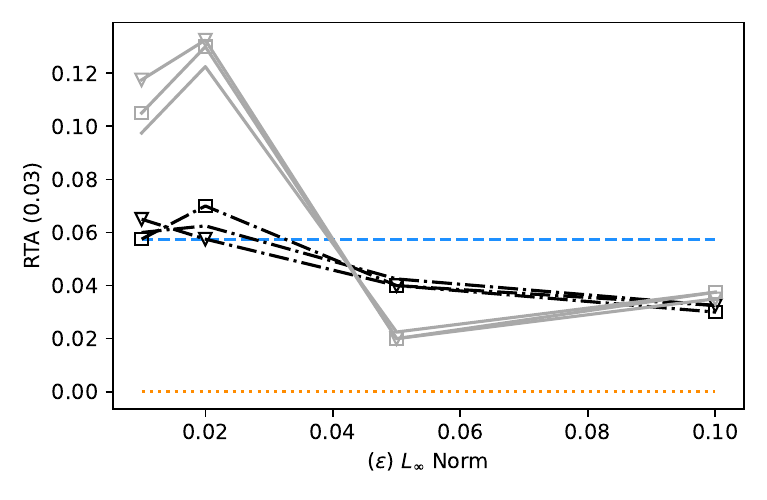}
        \caption{CIFAR-10}
    \end{subfigure}
    \begin{subfigure}{0.32\linewidth}
        \centering
        \includegraphics[page=2, width=\linewidth]{Figures/adversarial_trend0.02.pdf}
        \caption{Fashion-MNIST}
    \end{subfigure}
    \begin{subfigure}{0.32\linewidth}
        \centering
        \includegraphics[page=3, width=\linewidth]{Figures/adversarial_trend0.02.pdf}
        \caption{MNIST}
    \end{subfigure}
    \caption{Analyzing the effect of varying $\epsilon$ (see eq-\eqref{eq:amia_loss}) on the performance of AMIA and E-AMIA for the three datasets considered in this paper. \take{AMIA and E-AMIA are most effective at $\epsilon=0.02$}. \settings{$\sigma_n=0.02$, $N=50$, $\mathcal{I}$ in $LR_n$, $LR_p$ and $LR_o$}. }
    \label{fig:adv_trend}
\end{figure*}

\subsubsection{Effects of \tmath{$\epsilon$} on the attack performance}

Fig.~\ref{fig:adv_trend}(a)-(c) show the effects of varying $\epsilon$, the $l_\infty$ norm of the adversarial perturbations, on the performance of AMIA on CIFAR-10, Fashion-MNIST and MNIST datasets respectively. TPR of f-LiRA,$LR_f$ and EMIA,$LR_f$ are also provided for comparison. 

For all the three datasets and the indicator functions, we observe that the performance of AMIA initially increases as $\epsilon$ is increased from 0.01 to 0.02. This is because a greater $l_\infty$ norm allows AMIA to analyze and process more information about the local loss around $(x,y)$. TPRs of AMIA reach a maximum value at $\epsilon = 0.02$, and then start decreasing as $\epsilon$ increases. This decrease in TPR is because a too large $\epsilon$ only sub-optimally captures the local loss trends around $(x,y)$.

\begin{figure*}
    \centering
    \includegraphics[width=0.85\linewidth]{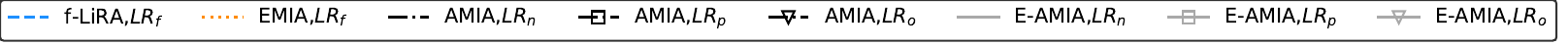}
    \begin{subfigure}{0.32\linewidth}
        \centering
        \includegraphics[page=1, width=\linewidth]{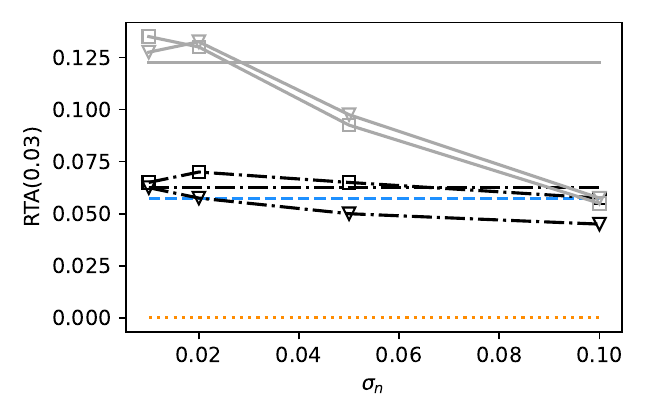}
        \caption{CIFAR-10}
    \end{subfigure}
    \begin{subfigure}{0.32\linewidth}
        \centering
        \includegraphics[page=2, width=\linewidth]{Figures/random_trend0.02.pdf}
        \caption{Fashion-MNIST}
    \end{subfigure}
    \begin{subfigure}{0.32\linewidth}
        \centering
        \includegraphics[page=3, width=\linewidth]{Figures/random_trend0.02.pdf}
        \caption{MNIST}
    \end{subfigure}
    
    \caption{Analyzing the effect of varying $\sigma_n$ (see eq-\eqref{eq:lr_p}) on the performance of AMIA and E-AMIA for the three datasets considered in this paper. \take{AMIA and E-AMIA are most effective at $\epsilon=0.02$}. \settings{$\epsilon=0.02$, $N=50$, $\mathcal{I}$ in $LR_n$, $LR_p$ and $LR_o$}. }
    \label{fig:random_trend}
\end{figure*}

\subsubsection{Effects of \tmath{$\sigma_n$} on the attack performance}

Fig.~\ref{fig:random_trend}(a)-(c) show the effects of varying $\sigma_n$, the standard deviation of Gaussian noise in eq-\eqref{eq:lr_p}, on the performance of AMIA on CIFAR-10, Fashion-MNIST and MNIST datasets respectively. TPR of f-LiRA,$LR_f$ and EMIA,$LR_f$ are also provided for comparison. 

For all the three datasets and the indicator functions, we observe that the performance of AMIA typically increases slightly as $\sigma_n$ is increased from 0.01 to 0.02, and then starts decreasing as $\sigma_n$ increases. The observed trend is interestingly similar to that observed when $\epsilon$ is increased in Fig.~\ref{fig:adv_trend}. We attribute this to the same reasons as above---the initial slight increase is caused by the greater $\sigma_n$ better capturing the local loss around $(x,y)$, and a too high $\sigma_n$ sub-optimally captures the local loss around $(x,y)$ causing a decrease in the running average.

However, interestingly, the decrease in the RTA due to an increase in $\sigma_n$ is significantly less drastic than that caused by an increase in $\epsilon$. We hypothesize that because $\sigma_n$ is the standard deviation of the Guassian noise, $n^l \sim \mathcal{N}(0, \sigma_n)$ is not bound by $l_\infty(\sigma_n)$, unlike the adversarial noise which is bound by $l_\infty(\epsilon)$. Therefore even if $\sigma_n$ is large, several noise elements have notably smaller $l_\infty$ norm due to the Guassian distribution. Therefore, a suboptimally large value of $\sigma_n$ better captures local loss trends than the suboptimally large value of $\epsilon$.

\begin{figure*}
    \centering
    \includegraphics[width=0.85\linewidth]{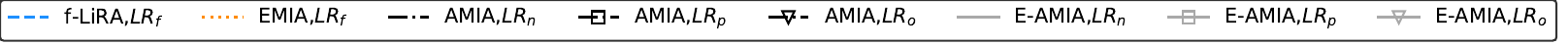}
    \begin{subfigure}{0.32\linewidth}
        \centering
        \includegraphics[page=1, width=\linewidth]{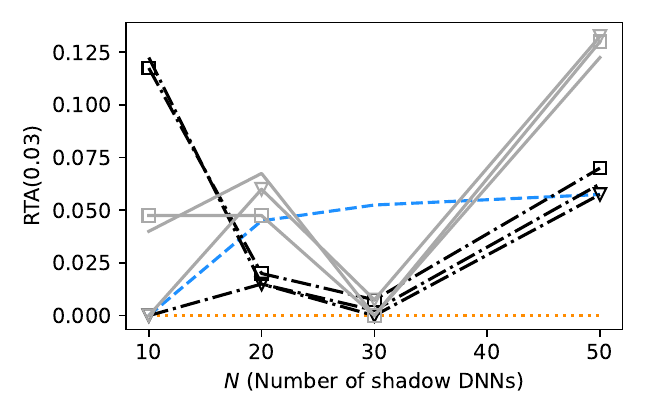}
        \caption{CIFAR-10}
    \end{subfigure}
    \begin{subfigure}{0.32\linewidth}
        \centering
        \includegraphics[page=2, width=\linewidth]{Figures/N_trend0.02.pdf}
        \caption{Fashion-MNIST}
    \end{subfigure}
    \begin{subfigure}{0.32\linewidth}
        \centering
        \includegraphics[page=3, width=\linewidth]{Figures/N_trend0.02.pdf}
        \caption{MNIST}
    \end{subfigure}
    
    \caption{Analyzing the effect of varying $N$ (see eq-\eqref{eq:amia_loss}) on the performance of AMIA and E-AMIA for the three datasets considered in this paper. \take{AMIA and E-AMIA are most effective at $\epsilon=0.02$}. \settings{$\epsilon=0.02$, $\sigma_n=0.02$, $\mathcal{I}$ in $LR_n$, $LR_p$ and $LR_o$}. }
    \label{fig:n_trend}
\end{figure*}

\subsubsection{Effects of \tmath{$N$} on the attack performance}
\label{sec:results_hyperparameters_N}

Fig.~\ref{fig:n_trend}(a)-(c) show the effects of varying $N$, the number of shadow DNNs trained by the attacker on the sampled datasets $D^j_a$, on the performance of AMIA on CIFAR-10, Fashion-MNIST and MNIST datasets respectively. TPR of f-LiRA,$LR_f$ and EMIA,$LR_p$ are also provided for comparison. We observe that the effectiveness of the attacks increases as $N$ is increased. In MI attacks, shadow DNNs are used to by the attacker to study the behavior of DNN on the subject given that the subject is or is not part of its training set. Therefore, training a greater number of shadow DNNs enables the attacker to more precisely approximate the expected behavior of the target DNN, ultimately improving the attack performance.

It can be observed in the figure that AMIA outperforms EMIA,$LR_f$ on all three considered datasets. On CIFAR-10 dataset, f-LiRA,$LR_f$ outperforms AMIA with a significant margin, except at $N=50$ where AMIA performs slightly better than f-LiRA,$LR_f$. We believe the reason for the increased effectiveness of f-LiRA,$LR_f$ on CIFAR-10 is the overfitting of the target CNN on the CIFAR-10 training set $D_t$. On Fashion-MNIST and MNIST datasets, where the target CNN is more generalized over the test set, AMIA is typically more effective than both f-LiRA,$LR_f$. For MNIST, $N=20$, f-LiRA,$LR_f$ shows surprisingly improved effectiveness---we regard it as an outlier, as we were unable to achieve a similar effectiveness with f-LiRA,$LR_f$ when we re-run the experiment.


\section{Discussion}

\begin{figure*}[!t]
    \centering
    \begin{subfigure}{0.32\linewidth}
        \centering
        \includegraphics[page=1, width=\linewidth]{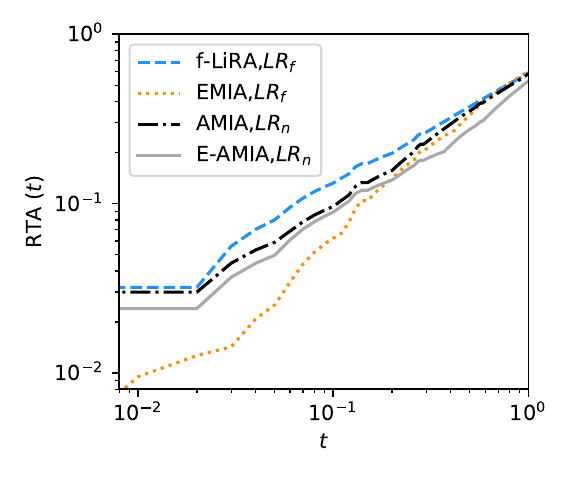}
        \caption{CIFAR-10}
    \end{subfigure}
    \begin{subfigure}{0.32\linewidth}
        \centering
        \includegraphics[page=2, width=\linewidth]{Figures/running_average_avg0.02,0.02.pdf}
        \caption{Fashion-MNIST}
    \end{subfigure}
    \begin{subfigure}{0.32\linewidth}
        \centering
        \includegraphics[page=3, width=\linewidth]{Figures/running_average_avg0.02,0.02.pdf}
        \caption{MNIST}
    \end{subfigure}
    
    \caption{Comparing the RTA of f-LiRA,$LR_f$ and EMIA,$LR_f$ with AMIA,$LR_n$ and E-AMIA,$LR_n$ as $t$ increases logarithmically when $\mathcal{V}$ (prepared for the target DNN) is transferred to unknown CNNs. Each is an average value computed over 10 unknown CNNs. \take{AMIA,$LR_n$ gives best performance overall, while both AMIA,$LR_n$ and E-AMIA,$LR_n$ perform better than f-LiRA,$LR_f$ and EMIA,$LR_f$ in low FPR region for Fashion-MNIST and MNIST datasets.} \settings{$\epsilon=0.02$, $N=50$.}. }
    \label{fig:transferability}
\end{figure*}
\subsection{Transferability of MI Attacks}
We study the transferability of MI attacks in Fig.~\ref{fig:transferability}(a)-(c) respectively for CIFAR-10, Fashion-MNIST and MNIST datasets. We define transferability as the ability of MI attack variables optimized for a target CNN to transfer to the unknown CNNs trained on the same dataset that the variables have not been optimized for. Fig.~\ref{fig:transferability} reports RTA of different attacks averaged over 10 unknown CNNs. We observe that generally in the low FPR region AMIA,$LR_f$ most effectively transfers to unknown CNNs, with E-AMIA,$LR_n$ and f-LiRA,$LR_f$ showing comparable performance. On the other hand, EMIA,$LR_f$ is least transferable to unknown CNNs even in the high FPR region where it performs best on the target CNN. This is because EMIA,$LR_f$ is more customized to the target DNN because of using soft labels. For the CIFAR-10 dataset however, f-LiRA,$LR_f$ outshines both AMIA,$LR_f$ and E-AMIA,$LR_n$ due to the overfitting of the unknown CNNs as discussed in detail in Section~\ref{sec:results_hyperparameters_N}.

Not surprisingly, the transferability of E-AMIA,$LR_n$ seems correlated with the transferability of EMIA,$LR_f$. More specifically, when EMIA,$LR_f$ shows worse transferability than AMIA,$LR_f$---for example on CIFAR-10 and Fashion-MNIST datasets in Fig.~\ref{fig:transferability}(a),(b)---E-AMIA,$LR_n$ consistently performs slightly worse than AMIA,$LR_f$. Likewise, when EMIA,$LR_f$ shows better transferability than AMIA,$LR_f$---for example on MNIST dataset in high TPR region---E-AMIA,$LR_n$ performs slightly better than AMIA,$LR_f$. This is simply because E-AMIA,$LR_n$ algorithm uses the data annotation approach used by EMIA,$LR_f$ to modify AMIA,$LR_f$.

\begin{figure}
    \centering
    \begin{subfigure}{0.49\linewidth}
        \centering
        \includegraphics[page=1, width=\linewidth]{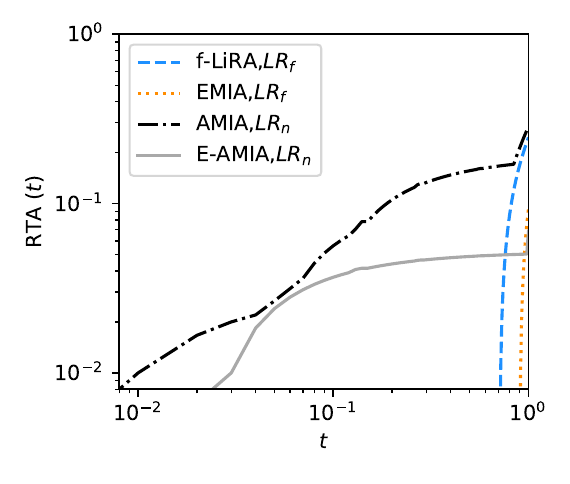}
        \caption{Fashion-MNIST}
    \end{subfigure}
    \begin{subfigure}{0.49\linewidth}
        \centering
        \includegraphics[page=2, width=\linewidth]{Figures/running_average_avg0.02,0.02_cnn_mini.pdf}
        \caption{MNIST}
    \end{subfigure}
    \caption{Comparing the RTA of f-LiRA,$LR_f$ and EMIA,$LR_f$ with AMIA,$LR_n$ and E-AMIA,$LR_n$ when $\mathcal{V}$ prepared for the target DNN is transfered to the unknown CNNs (that $\mathcal{V}$ has not been optimized for) trained with DP-SGD gradient updates as $t$ increases. Each is an average value computed over 10 unknown CNNs. \take{AMIA,$LR_n$ and E-AMIA,$LR_n$ outperform both f-LiRA,$LR_n$ and EMIA,$LR_n$ with a significant margin in low FPR region.} \settings{$\epsilon=0.02$, $N=50$}. }
    \label{fig:dp_sgd}
\end{figure}
\subsection{Evaluating Differentially Private Training}
Fig~\ref{fig:dp_sgd}(a)-(b) compares the transferability of f-LiRA,$LR_f$ and EMIA,$LR_f$ with that of AMIA,$LR_n$ and E-AMIA,$LR_n$ to the unknown CNNs trained with differential privacy algorithm (instead of with $L_2$ regularization as in Fig.~\ref{fig:transferability}) for Fashion-MNIST and MNIST datasets respectively. Due to large computational overhead, we refrain from optimizing new $\mathcal{V}$ directly over DP-SGD CNNs and reuse $\mathcal{V}$ already prepared for the target CNN in previous experiments. More specifically, each attack reuses $\mathcal{V}$, prepared for the target CNN in previous experiments, to infer the membership of given subjects for 10 unknown CNNs (that $\mathcal{V}$ has not been optimized for) trained with $(e=1.56, \delta=10^{-5})$-DP-SGD gradient updates.

We make two key observations. (1) CNNs trained with DP-SGD are notably more robust to MI attacks as compared to those trained with $L_2$ regularization as illustrated by the reduced RTA in Fig.~\ref{fig:dp_sgd}. This observation is consistent with that of Choquette et al.~\cite{choquette2021label} who note that only $L_2$ regularization needs to be notably stronger in order to match the robustness of DP-SGD to privacy attacks. (2) AMIA,$LR_n$ and E-AMIA,$LR_n$ are significantly more effective against DP-SGD as compared to f-LiRA,$LR_f$ and EMIA,$LR_f$. This is because (a) AMIA,$LR_n$ and E-AMIA,$LR_n$ are more transferable to unknown CNNs as compared to f-LiRA,$LR_f$ and EMIA,$LR_f$, and (b) AMIA,$LR_n$ and E-AMIA,$LR_n$ are significantly more effective against well-performing DNNs that do not overfit their training data.

\subsection{Limitations and Future Work}
AMIA, similar to LiRA and EMIA, assumes access to the real data distribution $D_r$, from which the training data $D_t$ is sampled. This assumption is commonly shared by several standard MI attacks. Although it might be challenging to meet this assumption in certain real-world scenarios, there exist other (more complex) approaches, such as the model inversion and the model stealing attacks, that can be used to estimate the training dataset $D_t$, given that $f_t$ is vulnerable to MI attacks under the aforementioned assumption. For example, an attacker may formulate the optimization problem to maximize $\mathcal{A}(\cdot)$ over the subject (similar to model inversion attacks~\cite{fredrikson2015model, zhang2020secret}). Therefore, MI attack is believed to be a fundamental privacy attack, and understanding the vulnerabilities of $f_t$ against MI attack is both critical and insightful~\cite{carlini2022membership}.


While our proposed attacks, AMIA and E-AMIA, to the best of our knowledge, are currently the most computationally efficient methods for utilizing subject membership information, they still require training multiple shadow DNNs. To further enhance efficiency, we suggest incorporating universal adversarial perturbations \cite{moosavi2017universal} that optimize the loss of member shadow DNNs and maximize the loss of non-member shadow DNNs. Our future work aims to develop a universal adversarial membership inference attack (U-AMIA) building upon this approach.

\section{Conclusions}
We first present a unified framework of current MI attacks divided it into three stages---preparation, indication and decision. We note several limitations of Likelihood Ratio Attack (LiRA) and Enhanced Membership Inference Attack (EMIA) at the preparation stage. Online LiRA is computationally highly inefficient, while offline LiRA and EMIA sacrifice the membership information of subjects for computational efficiency. This makes LiRA and EMIA perform poorly on MNIST and Fashion-MNIST datasets where the target is well-trained, unlike complex datasets (CIFAR-10, CIFAR-100, and Imagenet) where the target DNN typically overfits its training set due to the experimental setup of MI attacks. 
To address this, we propose Adversarial Membership Inference Attack (AMIA) that efficiently utilizes the membership information by training member shadow DNNs on subject batches. AMIA also optimally leverages the loss landscape around subjects by computing adversarially minimizing a novel loss function. We also propose the Enhanced AMIA (E-AMIA) that combines EMIA and AMIA to further improve the attack effectiveness. We experiment with a range of hyperparameters and observe that E-AMIA and AMIA notably outperform both LiRA and EMIA in the low FPR region---e.g., LiRA and EMIA showed 0\% TPR at 1\% FPR (99\% confidence), while AMIA and E-AMIA showed 6\% and 8\% TPR on Fashion-MNIST. We also study the transferability of each MI attack and show that AMIA is most transferable followed by E-AMIA and LiRA, while EMIA is least transferable.

\section*{Acknowledgment}
This publication was made possible by NPRP grant \# [13S-0206-200273] from the Qatar National Research Fund (a member of Qatar Foundation). The statements made herein are solely the responsibility of the authors.

\bibliographystyle{2_IEEE_Transactions/style}
\bibliography{bibliography/biblio1, bibliography/biblio2, bibliography/biblio3}










\end{document}